\documentclass[letterpaper, 10 pt, conference]{ieeeconf}  

\IEEEoverridecommandlockouts                              

\usepackage{graphics} 
\usepackage{epsfig} 
\usepackage{cite}
\usepackage{subfigure}
\usepackage{amsmath}
\usepackage{textcomp}
\usepackage{hyperref}
\usepackage{flushend}

\title{\LARGE \bf
Drone Detection Using Depth Maps
}

\author{Adrian Carrio$^{1}$, Sai Vemprala$^{2}$, Andres Ripoll$^{1,3}$, Srikanth Saripalli$^{2}$ and Pascual Campoy$^{1,3}$
\thanks{$^{1}$Adrian Carrio, Andres Ripoll and Pascual Campoy are with the Computer Vision \& Aerial Robotics Group, Universidad Polit\'ecnica de Madrid (UPM-CSIC), Madrid, 28003, Spain.
        {\tt\small adrian.carrio@upm.es}}%
\thanks{$^{2}$Sai Vemprala and Srikanth Saripalli are with the Department of Mechanical Engineering, Texas A\&M University, College Station, TX 77843, USA}%
\thanks{$^{3}$Andres Ripoll and Pascual Campoy are with the Delft Center for Systems and Control	(DCSC), Technische Universiteit Delft (TU Delft), Mekelweg 2, 2628 CD Delft, The Netherlands}%
}

\begin{document}

\maketitle
\thispagestyle{empty}
\pagestyle{empty}

\begin{abstract}

Obstacle avoidance is a key feature for safe Unmanned Aerial Vehicle (UAV) navigation. While solutions have been proposed for static
obstacle avoidance, systems enabling avoidance of dynamic 
objects, such as drones, are hard to implement
due to the detection range and field-of-view (FOV) requirements,
as well as the constraints for integrating such systems
on-board small UAVs. In this work, a dataset of 6k
synthetic depth maps of drones has been generated and used
to train a state-of-the-art deep learning-based drone detection model. While many sensing technologies can only provide relative altitude and azimuth of an obstacle, our depth map-based approach enables full 3D localization of the obstacle. This is extremely useful for collision avoidance, as 3D localization of detected drones is key to perform
efficient collision-free path planning. The proposed detection technique has been validated in several real depth map sequences, with multiple types of drones flying at up to 2 m/s, achieving an average precision of 98.7\%, an average recall of 74.7\% and a record detection range of 9.5 meters.

\end{abstract}

\section{INTRODUCTION}

Unmanned aerial vehicles (UAVs), or drones, are a popular choice for robotic applications given their advantages such as small size, agility and ability to navigate through remote or cluttered environments. Drones are currently being widely used for surveying, mapping with many more applications being researched such as reconnaissance, disaster management, etc. and therefore, the ability of a system to detect drones has multiple applications. Such technologies can be deployed in security systems to prevent drone attacks in critical infrastructures (e.g. government buildings, nuclear plants) or to provide enhanced security in large scale venues, such as stadiums. At the same time, this technology can be used on-board drones themselves to avoid drone-to-drone collisions. As an exteroceptive sensing mechanism, electro-optical sensors provide a small, passive, low-cost and low-weight solution for drone detection and are therefore suitable for this specific application. Additionally, drone detection typically requires large detection ranges and wide FOVs, as they provide more time for effective reaction.

In the literature, drone detection using image sensors has been proposed mainly in the visible spectrum \cite{7299040, 5938030, wu2017vision}. 
Thermal infrared imaging has also been proposed for drone detection \cite{ANDRASI2017183}. Thermal images typically have lower resolutions than those in visible spectrum cameras, but they have the advantage that they can operate at night.

Several other sensing technologies have been applied for drone detection (radar \cite{7497351} and other RF-based sensors \cite{Nguyen:2016:ICR:2935620.2935632}, acoustic sensors \cite{7382945} and LIDAR \cite{7778108}). Hybrid approaches have as well been proposed \cite{christnacher2016optical}. However, some of these technologies have limitations for being integrated on-board small drones, mainly their high power consumption, weight and size requirements and cost.

Image-based detection systems typically rely either on background subtraction methods \cite{ganti2016implementation}, or on the extraction of visual features, either manually, using morphological operations to extract background contrast features \cite{lai2011airborne} or automatically using deep learning methods \cite{8078541, 8078539}. Rozantsev et al. \cite{7299040} present a comparison between the performance of various of these methods. The aforementioned detection techniques rely on the assumption that there is enough contrast between the drone and the background. Depth maps, which can be obtained from different sensors (stereo cameras, \mbox{RGB-D} sensors or LIDAR), do not have these requirements.

\begin{figure*}
	\centering
	\subfigure[]{\includegraphics[width=81mm]{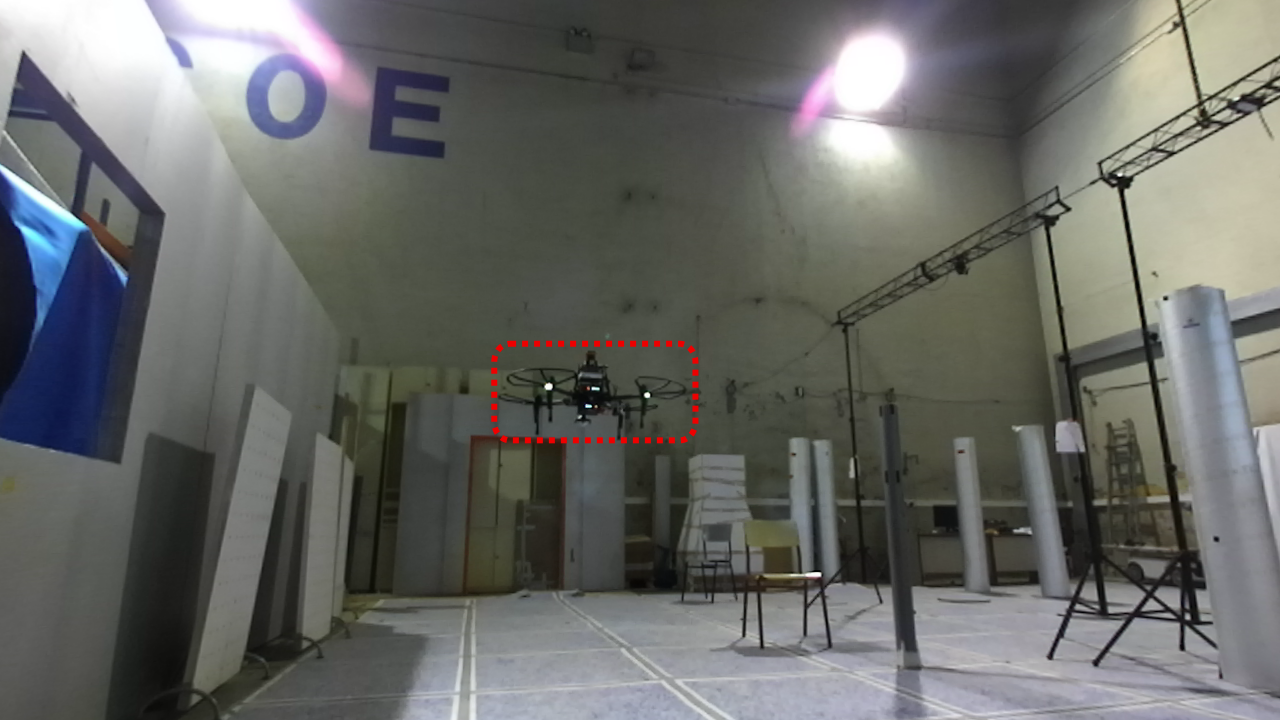}}
	\subfigure[]{\includegraphics[width=81mm]{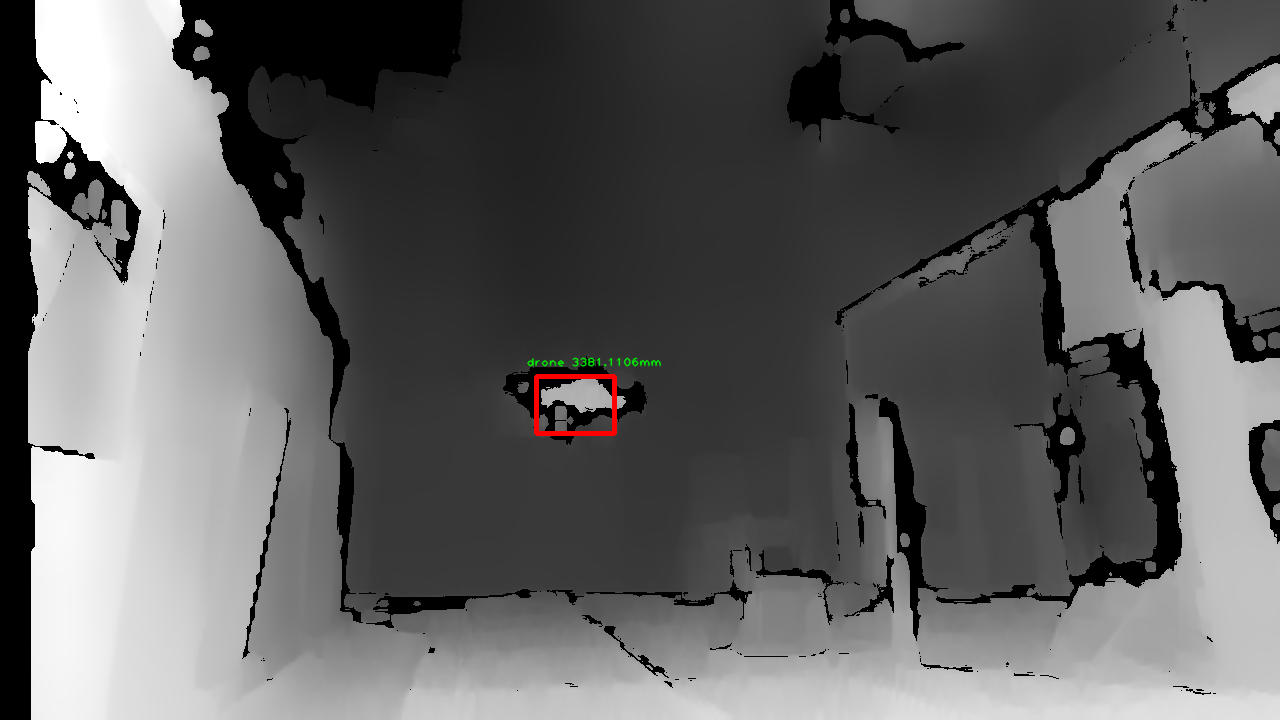}}
	\caption{The above images, RGB (left) and depth map (right), captured simultaneously, intuitively illustrate how the concept of depth contrast, as opposed to visual contrast, can be a better choice for drone detection. Based on this concept, we propose a novel, alternative method for drone detection only using depth maps.} 
	\label{fig:sample_images}
\end{figure*}

3D point clouds have been recently proposed for obstacle avoidance onboard drones using an RGB-D camera \cite{7989677}, but focusing on the detection of static obstacles. An alternative representation for point clouds are depth maps, which have been proposed for general object detection \cite{song2014sliding} and human detection \cite{xia2011human}, providing better detection performance as compared to RGB images. In the context of drone detection, a key concept that explains the usefulness of depth maps is that any flying object in a depth map appears with depth contrast with respect to the background. This happens as there are typically no objects with consistently the same depth around it. In other words, a flying object should generate a discontinuity in the depth map, which can be used as a distinct visual feature for drone detection. This concept is depicted in Fig. \ref{fig:sample_images}. An additional advantage of detecting using depth maps is that, while data from other sensing technologies can generally provide relative altitude and azimuth of the object only, depth maps can provide full 3D relative localization of the objects. This is particularly useful in the case of obstacle avoidance for drones, since the 3D position of the drone can be exploited to perform effective collision-free path planning. 

In this paper, we present a dataset of synthetic, annotated depth maps for drone detection. Furthermore, we propose a novel method for drone detection using deep neural networks, which relies only on depth maps and provides 3D localization of the detected drone. To the best of the authors' knowledge, this is the first time that depth maps are used for drone detection. The proposed detection method has been evaluated in a series of real experiments in which different types of drones fly towards a stereo camera. The reason to choose a stereo camera as the depth sensor for this work is the trade-off they provide in terms of detection range, FOV, lightweight and small size.

The remainder of this paper is as follows. Firstly, in section \ref{sec:detection_method}, we present our drone detection method. Secondly, in section \ref{sec:dataset}, details about the synthetic drone depth map dataset are presented. Thirdly, in section \ref{sec:implementation}, we describe the implementation details. In section \ref{sec:results}, we present the results of the proposed method and finally, in section \ref{sec:conclusions}, we present the conclusions and future work.

\section{DETECTION AND LOCALIZATION METHOD}
\label{sec:detection_method}

The proposed method for drone detection relies only on depth maps. Given a depth map, first, a trained deep neural network is used to predict the bounding boxes containing a drone and a confidence value for each bounding box. In order to localize the drone with respect to the camera, as the next step, a 2D point in each bounding box is chosen as actually belonging to the drone. The chosen point is then reprojected to 3D to get the actual drone relative position.

\subsection{Deep Neural Network}

\begin{figure}
\includegraphics[width=80mm]{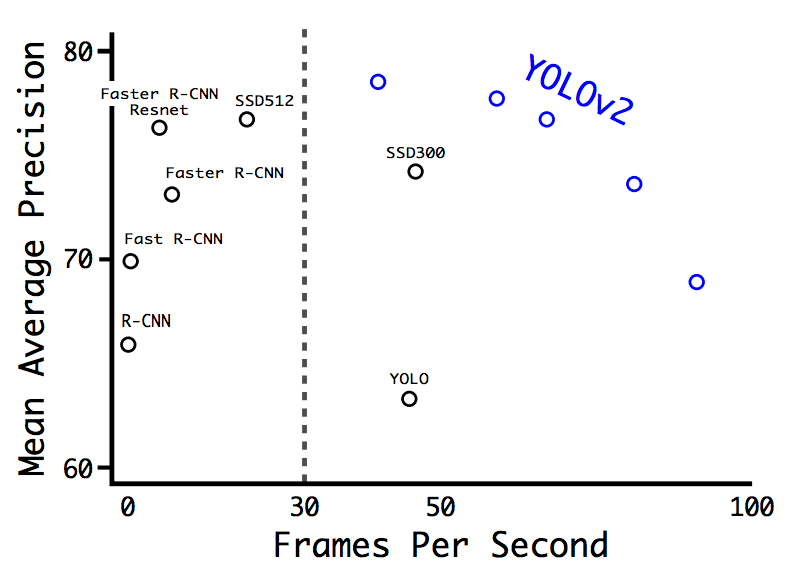}
\caption{Accuracy and speed of different object detection models on VOC 2007. The blue dots correspond to arbitrary input image sizes at which the YOLOv2 model can operate, even after it has been trained with a different input image size. In this way, the model provides a customizable trade-off between speed and accuracy.}
\label{fig:voc}
\end{figure}

YOLOv2 \cite{redmon2016yolo9000} is currently one of the fastest object detection algorithms, having obtained one of the highest performances in both speed and precision reported for the VOC 2007 challenge (see Fig.\ref{fig:voc}). It is also a very versatile model, as the input image size can be modified even after the model has been trained, allowing for an easy tradeoff between speed and precision. 

In YOLOv2, a single convolutional neural network predicts bounding boxes and class probabilities directly from full images in a single forward pass. This model has also been proposed for drone detection using RGB images \cite{8078539}.

\subsection{2D position of the drone in the depth image}

The bounding boxes predicted by the model do not always accurately indicate the actual position of the drone in the depth image. In the case of stereo vision, this happens mainly due to noise or errors in the stereo matching process. We propose the following method as a means to handle these potential inaccuracies.

Let $P=\{P_1,P_2,...P_n\}$ be the set of 2D points within the bounding box and $Z =\{Z_1,Z_2,...Z_n\}$ a set with their associated depths. We wish to choose a point $P_i \in P$ in the depth map which best indicates the position of the drone. We do this by choosing $P_i$ such that $i = argmin(|Z_i-Z_{ref}|)$. Let $Q1$ be the first quartile of $Z$.

Three different methods for choosing $Z_{ref}$ are proposed:
\begin{itemize}
\item Method 1 simply consists of choosing the 2D point with the minimum depth within the bounding box, or equivalently:

\begin{equation}
Z_{ref} = min(Z_i)
\end{equation}

\item Method 2 picks the 2D point with the closest depth to the mean of the 25\% smallest depths within the bounding box.
 
\begin{equation}
Z_{ref} = mean(Z_i) \forall Z_i<Q1
\end{equation}

\item Method 3 picks the 2D point with the closest depth to the median of the 25\% smallest depths within the bounding box.
 
\begin{equation}
Z_{ref} = median(Z_i) \forall Z_i<Q1
\end{equation}
\end{itemize}

In these methods, points that are further away are discarded, as the object to be detected should be closer to the camera than the background. Method 1 is the simplest, but also the most sensitive to spurious depth measurements as it relies on a single measurement. Methods 2 and 3 are intended to be robustified versions of method 1.

\subsection{3D localization}

In the case of a stereo camera, it is possible to estimate the 3D coordinates corresponding to the previously designated point $P_i(u,v)$ with disparity $d$ using Eq.~\ref{eq:reprojection}.

\begin{align}
   \begin{bmatrix} X \\ Y \\Z \end{bmatrix} &= \frac{T}{c_x^l-c_x^r-d} \begin{bmatrix}
           u-c_x^l \\
           v-c_y^l \\
           f^l
         \end{bmatrix}
         \label{eq:reprojection}
\end{align}

where $C^l(c_x^l,c_y^l)$ is the principal point  and $f^l$ is the focal length of the left camera and $C^r(c_x^r,c_y^r)$ is the principal point of the right camera.
\section{DATASET}
\label{sec:dataset}

In order to train a deep neural network for successful drone detection and evaluate it, we create a synthetic dataset of depth and segmentation maps for several sample drone platforms\footnote{The dataset can be found at: \url{https://vision4uav.com/Datasets}}. We utilize the UAV simulator Microsoft AirSim to construct simulated environments inside which drones are instantiated. Microsoft AirSim \cite{airsim2017fsr} is a recently released simulator for unmanned aerial vehicles which is built upon Unreal Engine: a popular videogame engine that provides capabilities such as high fidelity and high resolution textures, realistic post-processing, soft shadows, photometric lighting etc. These features make the combination of AirSim and Unreal Engine a particularly good choice for modeling cameras onboard drones and obtain the resultant images. Over the base functionality of Unreal Engine, AirSim provides flight models for drones as well as basic flight control features. 

\begin{figure}
	\centering
	\subfigure[]{\includegraphics[width=3cm,height=2cm]{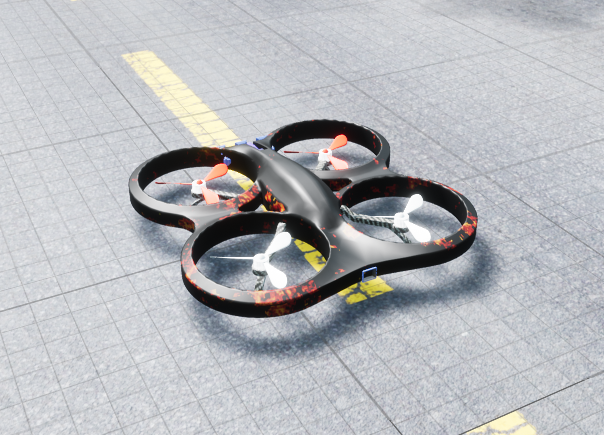}}
	\subfigure[]{\includegraphics[width=3cm,height=2cm]{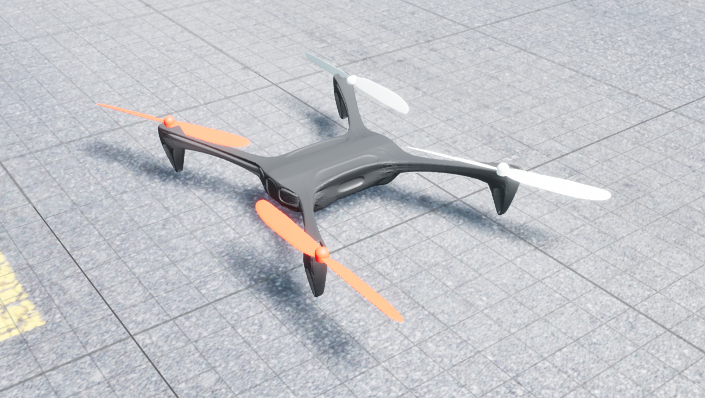}}
	\subfigure[]{\includegraphics[width=3cm,height=2cm]{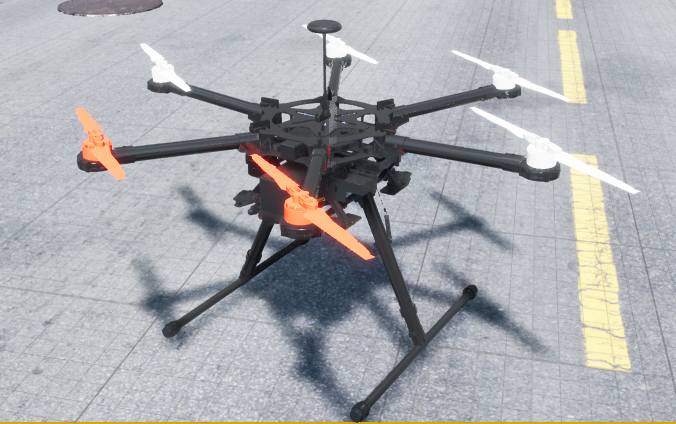}}
	\caption{Figures of the three drone models that are part of the training dataset. (a) Quadrotor, resembling a Parrot AR Drone. (b) Quadrotor, resembling a 3DR Solo. (c) Hexrotor, resembling a DJI S800.} 
	\label{fig:drone_models}
\end{figure}

\begin{figure}
	\centering
	\subfigure[]{\includegraphics[width=4.2cm,height=2.5cm]{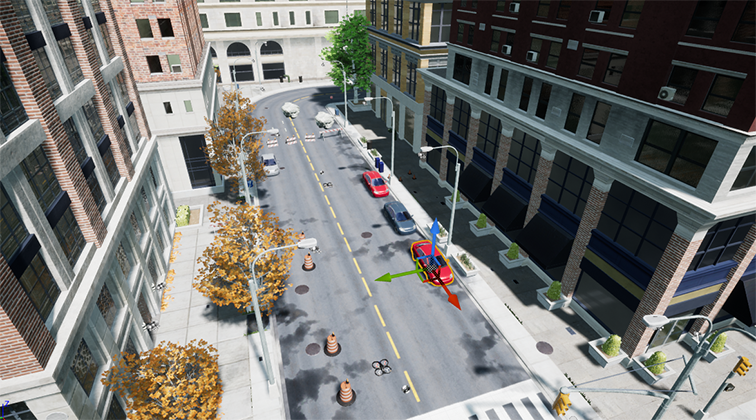}}
	\subfigure[]{\includegraphics[width=4.2cm,height=2.5cm]{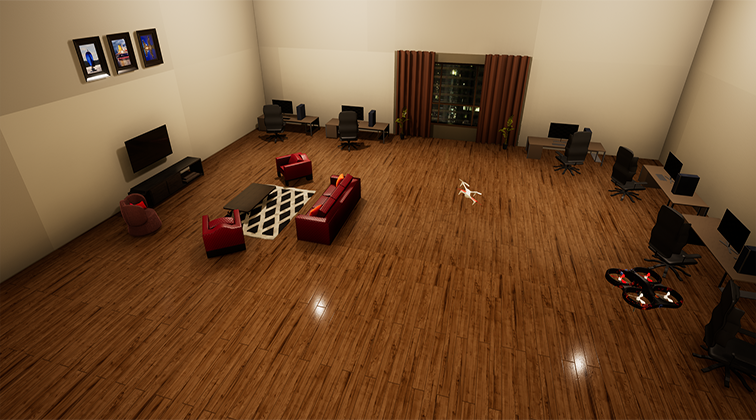}}
	\caption{Environments created within Unreal Engine simulate an urban outdoor environment (left) and an indoor environment (right), within which we instantiate multiple drones and obtain depth maps for training images} 
	\label{fig:environments}
\end{figure}

To create our synthetic dataset, we enhance the current functionality of AirSim by adding multiple models of drones. AirSim provides a base model for a quadrotor which resembles a Parrot AR Drone. For a more diverse representation of the appearance of drones, we create two additional models: one, a hexrotor platform resembling the DJI S800 and another quadrotor platform resembling a 3DR Solo. In Fig~\ref{fig:drone_models}, we show images of the three models used in AirSim that are included in the released dataset. AirSim contains an internal camera model, which we replicate to create stereo camera functionality for the drones. 
Through this functionality, we have generated more than 6k images of the three aforementioned types of drones, in various types of environments. For this purpose, we build and use custom environments within Unreal Engine. In particular, our dataset includes three different environments: an indoor office space environment, and outdoor environment with trees, buildings, etc. and a simple environment containing only a table with two chairs. In all the scenes, one of the drones is considered to be a `host', from which depth maps are obtained: and the other drone(s) that are visible in the depth maps are considered to be `target' drones, which are being observed. Figure \ref{fig:environments} shows pictures of the indoor and outdoor environments used.

\begin{figure*}
	\centering
	\subfigure[]{\includegraphics[width=54mm]{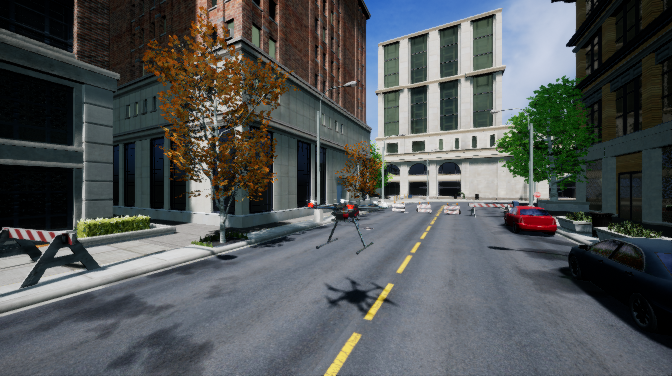}}
    \subfigure[]{\includegraphics[width=54mm]{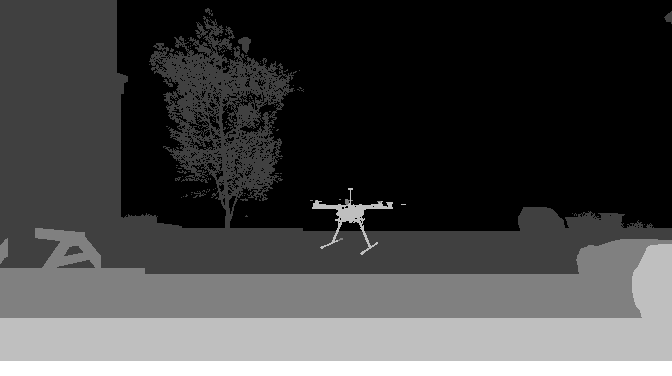}}
    \subfigure[]{\includegraphics[width=54mm]{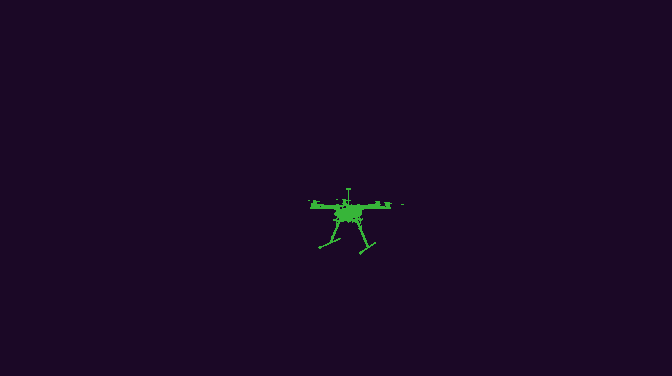}}
	\caption{Sample images from the dataset. In (a), the RGB image from the `host' drone's perspective is shown for reference, where it views a `target' drone, a hexrotor. The corresponding depth map is shown in (b), and (c) shows the segmentation image that isolates only the target drone.} 
	\label{fig:dataset_images}
\end{figure*}

In our dataset, we include at least two types of images. First, we render and record the disparity image obtained from the drone's viewpoint as per the preset baseline and resolution. Secondly, we include a segmentation image in which the drone(s) being observed is isolated. As Unreal Engine has the ability to keep track of all object materials in a scene, we identify only the materials that create the drone and isolate them to create the corresponding segmentation image. The location of the drone in the segmentation image is used later in order to create bounding boxes for the target drone, which are subsequently used for training the deep neural network. For the indoor and the outdoor environments we also include the RGB images. We record these images of the target drone from various distances, viewpoints and angles in three dimensions, attempting to simulate the observation of a drone hovering as well as in motion. In Fig.~\ref{fig:dataset_images}, we show sample depth and segmentation images generated from AirSim for the hexrotor model in an outdoor environment, along with the corresponding RGB image for reference.

\section{EXPERIMENTAL SETUP}
\label{sec:implementation}

\subsection{Hardware}

Once the deep neural network was trained with images from the synthetic depth map dataset, our experiments were aimed at using this model to detect real drones, assuming deployment onboard a mini UAV. Hardware for the experiments was selected trying to minimize size, weight and power demands, considering the limitations of applications onboard small drones.

The StereoLabs ZED stereo camera \cite{ZED} was selected as our imaging sensor due to its excellent features: high FOV (110$^\circ$ diagonal), low size and weight (175 x 30 x 33 mm, 159g) and acquisition speed (16 fps with HD1080 images). HD1080 video mode was selected in order to improve the detection of smaller/more distant objects. An NVIDIA Jetson TX2 module (85 grams) was used for the image acquisition and processing.

\subsection{Model and Inference}

For compatibility reasons with the ZED stereo camera API, Darkflow \cite{darkflow}, a python implementation of YOLO based in Tensorflow was used. By doing this, images can be acquired with the ZED camera and passed directly to the model for inference.

A smaller version of the YOLOv2 model called Tiny YOLOv2 was chosen to obtain faster performance. This model was reported to have 57.1\% mean average precision (mAP) in the VOC 2007 dataset running at 207 fps in a NVIDIA Titan X GPU. The model runs at 20 fps in a Jetson TX2. In our implementation, we modify the model configuration to perform single object detection and increase the input image size from its original value of 416x416 to 672x672, in other to improve the detection of smaller or more distant objects.

Input depth maps were codified as 8-bit, 3-channel images. For this, we downsample the resolution of the single-channel depth maps provided by the camera from 32-bit to 8-bit and store the same information in each of the three channels. This was done for simplicity of the implementation, as the objective was to explore the feasibility of drone detection with depth maps.

\section{RESULTS}
\label{sec:results}

\subsection{Training results}

From the dataset presented in Section \ref{sec:dataset}, a subset of 3263 images containing depth maps corresponding only to the Parrot AR Drone model were extracted. This was done in a effort to evaluate the generalization capability of the model, as it would be later evaluated on depth maps containing different drones.

The Tiny YOLOv2 model was trained on these images using a desktop computer equipped with an NVIDIA GTX 1080Ti. 80\% of the images were used for training and 20\% for validation. After 420k iterations (about 4-5 days) the model achieved a validation IOU of 86.41\% and a recall rate of 99.85\%.

\subsection{Precision and recall}

In order to obtain live measurements of the precision and recall of the model in real flights, a Parrot AR Drone and a DJI Matrice 100 were flown in an indoor space. The drones were manually flown at up to 2 m/s towards the camera, which was kept static. The live video stream obtained from the ZED camera was processed using a Jetson TX2 development board. The average processing time measured was about 200 ms per frame. 
For a drone flying at 2 m/s, this is equivalent to a detection every 0.4m, which should be acceptable for collision avoidance as long as the detector can also provide a large enough detection range. The low framerate is caused by the GPU being simultaneously used by the ZED camera for stereo matching and by Darkflow for inference of the detection model. An optimized version of the detection software is currently under development.

We use precision and recall as evaluation metrics for the detector. Precision here indicates the number of frames with correct drone detections with respect to the number of frames for which the model predicted a drone, while recall here indicates the number of frames with correct detections with respect to the number of frames containing drones.

The detection results using a threshold of 0.7 for the detection confidence are shown in Table \ref{tab:precision_recall}. The model successfully generalizes from AR Drone depth maps, on which it was trained, to depth maps generated by other types of drones. While processing a live stream of depth maps, it achieves an average precision of 98.7\% and an average recall of 74.7\%\footnote{A video showing some detection results can be found at the following link: \url{https://vimeo.com/259441646}}.

\begin{table}[] 
\centering
\caption{Precision and recall in online detection}
\label{tab:precision_recall}
\begin{tabular}{ c | c | c | c | c }	
\begin{tabular}{c}
Video \\ sequence
\end{tabular} & \begin{tabular}{c}
No. of \\ frames
\end{tabular} & \begin{tabular}{c}
Drone \\ model
\end{tabular}
 & \begin{tabular}{c}
Precision \\ (\%)
\end{tabular} & \begin{tabular}{c}
Recall \\ (\%)
\end{tabular}	\\ \hline
1 & 77 & AR Drone & 96.6 & 74.0 \\
2 & 48 & AR Drone & 95.3 & 85.4 \\
3 & 39 & AR Drone & 100.0 & 66.6 \\
4 & 33 & AR Drone & 100.0 & 66.6 \\
5 & 27 & AR Drone & 100.0 & 77.7 \\
6 & 64 & DJI Matrice & 100.0 & 67.1 \\
7 & 35 & DJI Matrice & 100.0 & 77.1 \\ \hline
\multicolumn{3}{r}{Averaged precision and recall} & 98.7 & 74.7 \\
    \hline          
\end{tabular}
\end{table} 



\subsection{Depth range}

For assessing the depth range and its reliability, frames were acquired with the camera in a static position while a Parrot AR Drone hovered at different distances, ranging from 1.5 to almost 10 m. For each of those hovering positions, 10 image detections were registered and the depth of the drone was measured using a laser ranger with $\pm$3 mm accuracy, which was recorded as the ground truth. The averaged depth error for those 10 detections was computed using each of the 3 methods proposed in Section \ref{sec:detection_method}. While the proposed method has been proven valid to detect the drone while flying at up to 2 m/s, here it was put in a hovering position only to enable accurate depth assessment and never to increase its observability. The results are shown in Table \ref{tab:depth}.

\begin{table}[] 
\centering
\caption{Comparison of depth estimation methods}
\label{tab:depth}
\begin{tabular}{ c | c | c | c }	
& \multicolumn{3}{c}{Averaged depth RMS error (mm)} \\
\begin{tabular}{c}
Hovering \\ distance (mm)
\end{tabular} & Method 1
 & Method 2 & Method 3	\\ \hline
1555 &56 	& 101	&89 \\
2303 & 235 &	315 &	171\\
2750 & 149 &	213	&184\\
3265 & 30 &	1436	& 1356\\
4513 & 151 &	118 	&126\\
5022 & 401 &	230	&    239 \\
5990 & 69 &	823	&    616\\
7618 & 292 &	147	& 139\\
8113 & 108 &	760	&  610\\
9510 & 254 &	937&	  1042\\
    \hline  
Average per method & 175 & 508 & 457\\          
\end{tabular}
\end{table} 

The best method is the one that assigns to the drone the 2D point with the minimum depth in the bounding box (i.e. Method 1). It appears to be robust enough for the application, with a maximum error of 401 mm. The failure of other methods can be explained by the fact that in many cases, the points belonging to the drone are less than 25\% of the points with depth in the bounding box. 

Accurate drone detections at a distance of up to 9510 mm have been achieved using this method. At this record distance, depth measurements using the minimum distance method had a minimum error of 143 mm and a maximum error of 388 mm. This depth range greatly exceeds the one recently reported for collision avoidance onboard small drones using point clouds in \cite{7989677}. In their work, a max indoor range of 4000 mm was obtained using the Intel\textregistered RealSense\textsuperscript{TM} R200 RGB-D sensor.

\section{CONCLUSIONS AND FUTURE WORK}
\label{sec:conclusions}

In this paper, a novel drone detection approach using depth maps has been successfully validated for obstacle avoidance. A rich dataset of 6k synthetic depth maps using 3 different drone models has been generated using AirSim and released to the public, in order to enable further exploration of this technique.

A subset of these depth maps, generated only using a model resembling an AR Drone, were used to train YOLOv2, a deep learning model for real-time object detection. Experiments in a real scenario show that the model achieves high precision and recall not only when detecting using depth maps from a real Parrot AR Drone, but also from a DJI Matrice 100. In other words, the model generalizes well for different types of drones.

An advantage of depth sensing versus other detection methods is that a depth map is able to provide 3D relative localization of the detected objects. This is particularly useful for collision avoidance onboard drones, as the localization of the drone can be useful for effective collision-free path planning. The quality of this localization method has been assessed through a series of depth measurements with a Parrot AR Drone hovering at different positions while it was being detected. A record max depth range of 9510 mm was achieved, with an average error of 254 mm. To the best of our knowledge, this is the first time that depth maps are proposed for drone detection and subsequent localization.

As for future work, our immediate objective is to test the system onboard a flying drone. A C++ implementation of the model and inference will be explored in order to increase the execution speed. Additionally, a multi-object tracking approach using joint probabilistic data association will be implemented to provide continuous, real-time detection.


\section*{ACKNOWLEDGMENT}

The authors would like to thank Hriday Bavle and \'Angel Luis Mart\'inez for their help with the hardware used in the experiments.

\bibliographystyle{IEEEtran}
\bibliography{IEEEabrv,references}

\end{document}